\documentclass[11pt,a4paper]{article}
\usepackage{coling2018}
\usepackage{times}
\usepackage{latexsym}
\usepackage{url}
\usepackage{graphicx}
\usepackage{multicol}
\usepackage{multirow}

\title{Event Coreference Resolution Using Neural Network Classifiers}

\author{Arun Pandian \\
  Carnegie Mellon University in Qatar\\
  Doha, Qatar  \\
  {\tt apandian@andrew.cmu.edu} \\\And
  Lamana Mulaffer\\
  Carnegie Mellon University in Qatar \\
  Doha, Qatar  \\
  {\tt fmulaffe@andrew.cmu.edu} \\
  \AND
  Kemal Oflazer \\
  Carnegie Mellon University in Qatar\\
  Doha, Qatar  \\
  {\tt ko@cs.cmu.edu} \\\And
  Amna AlZeyara\\
  Carnegie Mellon University in Qatar \\
  Doha, Qatar  \\
  {\tt azeyara@andrew.cmu.edu} \\
}

\date{}

\begin{document}

\maketitle
\begin{abstract}
This paper presents a neural network classifier approach to  detecting
both within- and cross-document event coreference effectively using
only event mention based features. Our approach does not (yet) rely on
any event argument features such as semantic roles or spatiotemporal
arguments.  Experimental results on the ECB+ dataset show that our
approach produces $F_1$ scores  that significantly outperform the state-of-the-art methods for both within-document  and cross-document event coreference resolution when we use $B^3$ and $CEAF_e$
evaluation measures, but  gets  worse $F_1$ score with the $MUC$
measure. However, when we use the $CoNLL$ measure,
which is the average of these three scores, our approach has slightly
 better $F_1$ for within-document event coreference resolution but is
 significantly better for cross-document event coreference
resolution.

\end{abstract}

\section{Introduction}
Event coreference resolution is the task of identifying spans of text that refer to unique events and clustering or chaining them, resulting in one cluster/chain per unique event. This is an important part of an NLP system that performs topic detection \cite{Allan-etal98}, information extraction, question answering \cite{NarayananandHarabagiu2004}, text summarization \cite{Azzam-etal1999} or any other system that is predicated on understanding natural language. Unlike entity coreference, which refers to clustering of nouns or pronouns that refer to the same entity, event coreference is a fundamentally harder problem. This is because the event as a semantic unit is structurally more complex to identify and resolve coreference for, as it has \emph{event arguments} such as participants and spatio-temporal information that could be distributed across the text \cite{BejanandHarabagiu2010}. Furthermore, different event mentions can refer to the same real world event and thus the context of the mentions and their arguments may also need to be considered \cite{Yang-etal2015}. For instance, in the sentences
\begin{quote}
  \begin{small}
\textit{Lindsay Lohan \textbf{checked into} New Promises Rehabilitation Facility on Sunday morning.}     
  \end{small}
\end{quote}
and 
\begin{quote}
  \begin{small}
\textit{News of this \textbf{development} caused the media to line up outside the facility.}    
  \end{small}
\end{quote}
``checked into'' and ``development'' are coreferent. However, devoid
of context,  ``checked into'' and ``development'' are not semantically related. Additionally, events arguments (e.g., Lindsay Lohan, New Promises Rehabilitation Facility, Sunday morning) don't appear in the second sentence. 

There are two types of event coreference resolution, within-document
(WD) and cross-document (CD) event resolution. WD resolution is
typically easier to solve as there is a higher chance of true
coreference if there is similarity in the words used, contexts and
event arguments. On the other hand, more evidence is needed to resolve
coreference across documents as different documents are less likely
to talk about the same events in the same way. Therefore, it is common
practice to first solve WD coreference and then later use this
information to solve CD coreference \cite{Choubey-etal2017}. In this
work, we try to solve both within-document and cross-document
coreference without identifying event arguments and their  semantic
roles in the event as these are still  difficult to extract with high accuracy \cite{BejanandHarabagiu2014}.

We build two feed-forward neural nets for  pairwise event coreference prediction, one for WD
coreference and one for CD coreference that output the probability
of coreference of the pair of events in question. Both classifiers are
trained separately since we expect that the importance for the
features for CD and WD coreference to differ \cite{Choubey-etal2017}.  

Once all pairwise event coreference predictions are complete, we
construct a graph where each node is an event mention and each edge
weight is the probability produced by the classifier representing a
potential coreference relation between the nodes. We then find all the
connected components (equivalent to finding the coreference clusters)
in the graph after edges are filtered by a predetermined threshold. In the case of CD coreference, we find connected components in two phases - first by finding all the WD connected components and second by merging these WD components to build CD components.

Experimental results on ECB+ dataset show that our
$F_1$ scores  significantly outperform the state-of-the-art methods for both
WD and CD event coreference resolution when we use $B^3$
\cite{BaggaandBaldwin1998} and $CEAF_e$ \cite{Luo2005} 
evaluation measures, but  gets  worse $F_1$ score with the $MUC$
measure \cite{Vilain-etal1995}. However, when we use the $CoNLL$ measure
which is the average of these three scores \cite{Pradhan-etal2014}, our approach has slightly
 better $F_1$ for WD resolution but is significantly better for CD
resolution.
\section{Related work}

Different approaches, focusing on either WD or CD coreference chains,
have been proposed for event coreference resolution. Works specific
to WD event coreference include pairwise classifiers
\cite{Ahn2006,Chen-etal2009}, graph-based clustering
\cite{ChenandJi2009} and information propagation
\cite{Liu-etal2014}. Works focusing purely on CD coreference include Cybulska and Vossen
\shortcite{CybulskaandVossen2015a} who created pairwise classifiers using
features indicating granularities of event slots, and in another work
\cite{CybulskaandVossen2015b}, use discourse analysis at the document
level along with `sentence' templates amongst documents that have
possibly coreferent events. Several papers have studied event
extraction and event coreference as a joint process \cite{araki2015joint,nglong2017}.

Several studies have considered both WD and CD event coreference
resolution tasks simultaneously. Such approaches
\cite{Lee-etal2012,BejanandHarabagiu2010,BejanandHarabagiu2014}
create a meta-document by concatenating topic-relevant documents and
treat both WD and CD coreference resolution as identical tasks. Most recently, Yang et
al. \shortcite{Yang-etal2015} applied a two-level clustering model
that first groups event mentions within a document and then groups WD
clusters across documents in a joint inference process. Choubey et
al. \shortcite{Choubey-etal2017} used an event coreference model that
uses both pairwise CD and WD classifiers to build event clusters
iteratively by switching between WD and CD coreference until the
results converge. While our approach also uses the two pairwise
classifiers, in our CD resolution model, we simply build WD components (clusters) followed by CD components (clusters) in one iteration.

\begin{figure*}[h]
  \begin{center}
  \includegraphics[width=\textwidth]{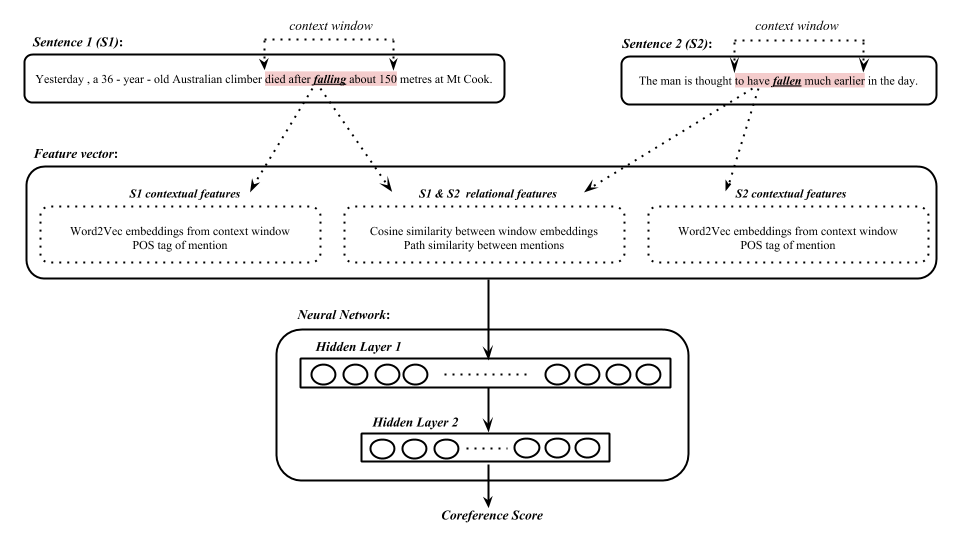}
  \end{center}
  \caption{Feed-forward Neural Network structure for CD Resolution}
  \label{cd_model}
\end{figure*}

\section{Detecting pairwise event coreference}

Both WD and CD event coreference resolution are implemented using feedforward neural nets
with ReLU units that take two featurized events, their contexts and
assorted semantic information,   and determine if the events are
coreferent. The WD NN is a single hidden layer neural net with 300
hidden units, while the CD NN has two hidden layers with 400 and 150
hidden units for layer 1 and layer 2 respectively. The size of the NNs
were determined by experiments documented in Section
\ref{Sec:Eval}. During each feedforward back-propagation cycle, the NN
aims to minimize the negative log cross entropy of the softmax
distribution.  The final output layer of both NNs is a softmax that
gives the probability of coreference between the pair of events.  

\subsection{Event features}
\label{feat}
We use two  types of features for characterizing event mentions:
\textit{contextual features} and \textit{relational
  features}. Contextual features are extracted from each sentence
independently and relational features depend on the relationships
between the two sentences.

The contextual features are the main event word, the POS tag of the
event word,  and words in a pre-defined window around the event word. In
representing the words in a sentence, we use word embeddings of size
400 from a word2vec model \cite{word2vec} that we built using the gensim
implementation from the English Wikipedia corpus
\cite{gensim}. The embeddings we included are that of the event mention
as well as the two words on each side that appear in the
word2vec model.\footnote{For multi-word events, we
  determined the head word using the approach by Honnibal and Johnson
  \shortcite{textblob} and used its embedding.} We generated the POS tags for the event mentions using
the Natural Language ToolKit (NLTK) package \cite{NLTKBook} and represented them as one-hot vectors. 

The relational features include the distance between the two sentences
(for WD classification only), the cosine similarity of the embeddings
of the event words and the WordNet  similarities using
variations of the main event words. The cosine similarity is
calculated for the embedding of the event words and the value is
quantized into 11 buckets (including one bucket for unknown similarities) and represented as  one-hot vectors. As for the WordNet
similarities, we consider three values using variations of the main
event words. All the path similarities, senses and derivation related
to WordNet are generated through TextBlob \cite{textblob}. We
calculate the first value by considering all the senses of both event
words. Different path similarities are calculated using two word
combinations; one word from each set of senses. Ultimately, the
maximum similarity is selected. We calculate the second value by using
the path similarity between the hypernyms of the event words. Lastly,
we calculate the third value by considering the path similarity
between derivationally related verb forms of each event word
provided by WordNet. Similar to the cosine similarity, all three
values are quantized and represented as a one-hot vector. Figure
~\ref{cd_model} shows an overview of the CD resolution model, which
takes as input a pair of sentences, featurizes them into neural
network inputs 
comprising of contextual and relational features, passes them through
the two layer NN and produces a coreference score, which is the
probability of coreference between the two event mentions in the
input sentences.

During the preliminary stage of feature selection, we started with
only contextual features; the event word embeddings, context embeddings and POS tags. Then we tested with only relational features;
distance between the sentences, word embeddings, cosine similarity and
WordNet path similarity of the main event words. Finally, we tested
with using both, contextual and relational features. The accuracy of
using both contextual and relational features was higher than when
using them separately as expected, but using either feature set
individually gave similar results on the development set.  

When we performed an error analysis, we discovered that despite an
observable semantic relation between words, the WordNet similarity was
low (especially for words that had the same hypernym).  Therefore, we quantized the 
WordNet path similarity between the
hypernyms and added it as a features as a one hot vector.  Additionally, we found that the WordNet path similarity
differentiated greatly between words of different syntactic
categories. Therefore, we also generated the derivationally related
verb form of each event word, and added the path similarity between
them. Adding WordNet path similarity of hypernyms and the related verb
form of each word greatly improved performance. Results further
improved when non-coreferent pairs where sampled purely from all pairs
whose mentions were either from the  same sentence or from
sentences that shared a coreference pair. 

\subsection{Changes to classifier from earlier work}

In comparison to the latest system that uses pairwise classifiers
\cite{Choubey-etal2017} which builds on \cite{Chen-etal2009,Ahn2006},
our pairwise classifiers have some significant differences. We use
precomputed word embeddings, as opposed to computing word embeddings
during classifier training. We also do not featurize event arguments
as part of our pairwise classifier (as mentioned above), as extracting
event arguments and their relation to an event is difficult to identify
accurately  \cite{BejanandHarabagiu2010}. Furthermore we use contextual features for both WD and CD coreference, while Choubey et al.,\shortcite{Choubey-etal2017} use contextual features only for CD coreference. The aforementioned system uses only cosine similarity and euclidean distance between their computed embeddings for their relational features, while we extract other relational features explained in \ref{feat}, including WD path similarities. 

\subsection{Training the pairwise coreference classifiers}
We train the pairwise classifiers using documents from topics 1-23 of
the ECB+ corpus \cite{Yang-etal2015,Choubey-etal2017}. The statistics of the corpus are provided in Table \ref{ecbstats}. We extract the
training data clusters, and generate all the coreferent event mention
pairs from them. To ensure the pairwise classifiers become proficient
at identifying non-coreferent event mentions even in similar
contexts, we only include those non-coreferent pairs whose event
mentions either belong to the same sentence or  belong to sentences
that also share a coreferent pair. We then sample from these
non-coreferent pairs to ensure the number of coreferent and
non-coreferent training pairs are the same. Although the number of non-coreferent pairs outweigh the number of coreferent pairs, we train our WD model using a 50-50 training split, that is equal number of coreferent and non-coreferent pairs, to avoid developing a bias for some statistical distribution of these pairs within the ECB+ corpus. However, we train our CD model using the actual training split, in order to provide more training samples as CD resolution is inherently a more difficult problem to solve than WD resolution. We train WD and CD pairwise classifiers separately since 
we expect the importance of neural net learned weights to differ between 
for the two cases \cite{Choubey-etal2017}. 

\begin{table}[t!]
\begin{center}
\begin{tabular}{|l|r|r|r|r|}
\hline
 & \textbf{Train} & \textbf{Dev} & \textbf{Test} & \textbf{Total} \\ \hline
\#Documents & 462 & 73 & 447 & 982 \\ \hline
\#Sentences & 7,294 & 649 & 7,867 & 15,810 \\ \hline
\#Event Mentions & 3,555 & 441 & 3,290 & 7,286 \\ \hline
\#CD Chains & 687 & 47 & 586 & 1,220 \\ \hline
\#WD Chains & 2,499 & 316 & 2,137 & 4,952 \\ \hline
  Avg. WD & 2.84 & 2.59 & 2.55 & 2.69 \\
   chain length & & & &\\\hline
Avg. CD  & 5.17 & 9.39 & 6.77 & 5.98 \\ chain length & & & & \\ \hline
\end{tabular}
\caption{ECB+ corpus statistics}
\label{ecbstats}
\end{center}
\end{table}

\subsection{Intrinsic Evaluation}
\label{Sec:Eval}
To ensure successful clustering, it was paramount to have a strong pairwise classifier. Therefore we performed intrinsic evaluations for this classifier. 
For this purpose, we use documents from topics (23-25) as the development set and
topics (26-45) as the test set \cite{Yang-etal2015,Choubey-etal2017}. The development set was used to tune the parameters of
the classifier as well as to find the probability threshold of the output that
determines coreference. 

As the ECB+ corpus is incompletely annotated in both event mentions and
event coreference \cite{cybulska2014guidelines}, running an
event detection tool will not be necessarily instructive as some
coreferences between events and actual events themselves are left
unmarked in the database. Therefore we extract gold standard event mentions. Regardless, to be able to compare our
results to previously published results 
\cite{Yang-etal2015,Choubey-etal2017}, we perform event detection
using the same event detection tool used by these systems on the same
test set. The event detection tool used is a CRF-based semi-Markov
model that is trained using sentences from ECB+ to provide more
accurate detection of events \cite{Yang-etal2015}. Once we extract the
event mentions, we only use those mentions also found in the gold
standard as the ECB+ corpus is incompletely marked and it's not
possible to determine if pairs generated from non-gold event mentions are
actually non-coreferent.

Once the mentions are finalized, for both detected event mentions and
gold standard event mentions, all possible coreferent pairs are
generated. For the WD case, the non-coreferent pairs are all pairs
within a document that are not marked coreferent, while for the CD
case, they comprise all pairs within a sub-topic that are not marked
coreferent.

As mentioned above, for  the first set of experiments we sample
non-coreference pairs equal to the number of coreference pairs and
evaluate our pairwise models on them.  For the second set of experiments we
used all the non-coreferent pairs without any sampling. We then used
our development set to determine the best coreference thresholds for
event clustering. We realized that for the purpose of finding event
clusters it was important to have high precision for the event
coreference. The best coreference thresholds were 0.95 for WD coreference
and 1.0 for CD coreference. Therefore we also reports results on the
these numbers for both the balanced and unbalanced test sets. For the
unbalanced test set, we report the results on both types of event
mention pairs, one generated from the gold standard event mentions
(called WD-gold and CD-gold) and one from  the extracted event
mentions (WD-detect and CD-detect).   

\begin{figure*}[h]
  \begin{center}
  \includegraphics[width=\textwidth]{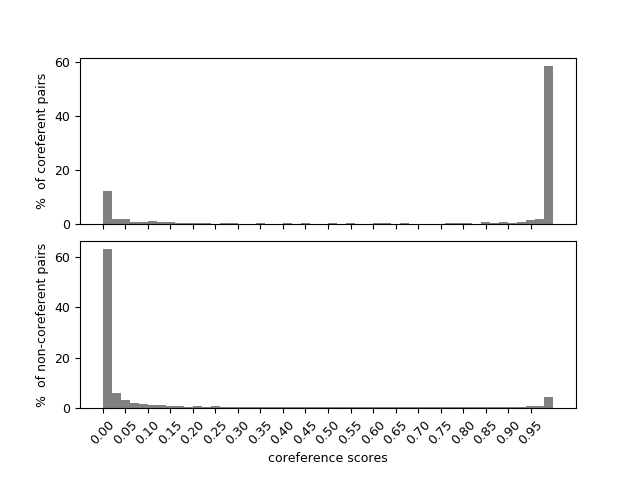}
  \end{center}
  \caption{Coreference score distribution from WD-gold model. Scores of coreferent pairs (top) and scores of non-coreferent pairs (bottom)}
  \label{histogram}
\end{figure*}

The results of experiments on the balanced test set are reported in Table \ref{pw_balanced} and of the whole test set in Table \ref{pw_unbalanced}. We see that the precision improves when the threshold increases in all cases. We also report the accuracy which represents the number of pairs classified correctly by our classifier. Additionally, the histogram in Figure~\ref{histogram}
shows the coreference score distribution from the WD resolution model
for coreferent and non-coreferent pairs of mentions. The CD resolution
model has a similar coreference score distribution. As we can see in
both Table \ref{pw_unbalanced} and  Figure \ref{histogram}, our system
is able to accurately detect and differentiate between coreferent
pairs quite competently. 

\begin{table*}[t!]
\begin{center}
\resizebox{\linewidth}{!}{%
\begin{tabular}{|l|r|r|r|r|r|r|r|r|r|}
\hline
 & \multicolumn{1}{c|}{Coreference } & \multicolumn{1}{c|}{\#Coref} & \multicolumn{1}{c|}{\#Non-coref} & \multicolumn{1}{c|}{TP} & \multicolumn{1}{c|}{TN} & \multicolumn{1}{c|}{P} & \multicolumn{1}{c|}{R} & \multicolumn{1}{c|}{$F_1$} & \multicolumn{1}{c|}{Accuracy} \\
 & \multicolumn{1}{c|}{Threshold} & \multicolumn{1}{c|}{Links} & \multicolumn{1}{c|}{Links} &  &  & &  &  &  \\ \hline 
WD-gold & 0.5 & 1,799 & 1,799 & 1,302 & 1,565 & 84.77 & 72.37 & 78.08 & 79.68 \\ \hline
WD-gold & 0.95 & 1,799 & 1,799 & 1,097 & 1,697 & 91.49 & 60.98 & 73.18 & 77.65 \\ \hline
CD-gold & 0.5 & 24,315 & 24,315 & 16,968 & 21,124 & 84.17 & 69.78 & 76.30 & 78.33 \\ \hline
CD-gold & 1 & 24,315 & 24,315 & 9,817 & 23,689 & 94.01 & 40.37 & 56.48 & 68.90 \\ \hline
\end{tabular}
}
\caption{Intrinsic evaluation for pairwise classifier with balanced test sets}
\label{pw_balanced}
\end{center}
\end{table*}

\begin{table*}[t!]
\begin{center}
\resizebox{\linewidth}{!}{%
\begin{tabular}{|l|r|r|r|r|r|r|r|r|r|r|}
\hline
 & \multicolumn{1}{c|}{Coreference} & \multicolumn{1}{c|}{\#Coref} & \multicolumn{1}{c|}{\#Non-coref} & \multicolumn{1}{c|}{TP} & \multicolumn{1}{c|}{TN} & \multicolumn{1}{c|}{P} & \multicolumn{1}{c|}{R} & \multicolumn{1}{c|}{$F_1$} & \multicolumn{1}{c|}{Accuracy} & \multicolumn{1}{c|}{Accuracy} \\ 
 & \multicolumn{1}{c|}{Threshold} & \multicolumn{1}{c|}{Links} & \multicolumn{1}{c|}{Links} &  &  &  &  &  & \multicolumn{1}{c|}{(Non-coref)} & \multicolumn{1}{c|}{(All)} \\ 
 \hline
WD-gold & 0.95 & 1,799 & 12,701 & 1,097 & 11,984 & 60.47 & 60.98 & 60.72 & 94.35 & 90.21 \\ \hline
WD-detect & 0.95 & 1,212 & 11,735 & 762 & 11,353 & 66.61 & 62.87 & 64.69 & 96.74 & 93.57 \\ \hline
CD-gold & 1.0 & 24,315 & 144,515 & 9,669 & 143,002 & 86.47 & 39.77 & 54.48 & 98.95 & 90.43 \\ \hline
CD-detect & 1.0 & 16,329 & 133,991 & 7,110 & 133,048 & 88.29 & 43.54 & 58.32 & 99.3 & 93.24 \\ \hline
\end{tabular}
}
\caption{Intrinsic evaluation for pairwise classifier with actual (unbalanced) test sets}
\label{pw_unbalanced}
\end{center}
\end{table*}

\section{Clustering Events}

From our pairwise classifier, our next step is to build clusters of event mentions such that all mentions in a cluster are considered coreferent to each other. In order to build such coreferent clusters, we model our problem as finding connected components in a weighted graph. We represent event mentions as nodes and coreference scores of event mention pairs (given by the pairwise classifier) as weights of the edges between those mention pairs. In
the case of WD resolution, an edge between a pair of mentions (nodes) exists
if they belong to the same document. In the case of CD resolution, an
edge between a pair of mentions (nodes) exists if they belong to the same
topic, since we know that there are no CD coreferences between event
mentions from different topics. 

For WD resolution, we filter out all edges with weights less than 0.95 and find all the connected components in the graph. These components are the WD coreferent clusters which we later evaluate against the WD gold standard clusters. For CD resolution, we first perform WD resolution on all WD edges. We then build CD components if there is a CD edge with a threshold of 1.0 between the WD components. These components are the CD coreferent clusters which we later evaluate against the CD gold standard clusters.

\section{Evaluation}

We perform all our experiments on the ECB+ news corpus \cite{CybulskaandVossen2014}. As described in \ref{ecbstats}, our test set consists of documents from topics 26-45. 

We evaluated our system using three widely used coreference resolution
metrics: $MUC$, $B^3$ and $CEAF_e$, computed using the most recent version
(v8.01) of the official CoNLL Scorer \cite{Pradhan-etal2014}. $MUC$
\cite{Vilain-etal1995} measures how many gold (predicted) cluster
merging operations are needed to recover each predicted (gold)
cluster. $B^3$ \cite{BaggaandBaldwin1998} measures the proportion of
overlap between the predicted and gold clusters for each mention and
computes the average scores. $CEAF_e$ \cite{Luo2005}  measures the best
alignment of the gold-standard and predicted clusters. We also
calculate the $CoNLL$ $F_1$ score, which is the average of the $F_1$ scores
across all three evaluation metrics.

\subsection{Baseline Systems}
We compare our system using three baseline systems.
\begin{enumerate}
\item LEMMA:  This baseline groups event mentions into clusters based on their lemmatized head words. A cluster is formed if all its event mentions share the same lemmatized head word. This is often considered a strong baseline \cite{Yang-etal2015}.
\item HDDCRP: This baseline is produced by the supervised Hierarchical Distance Dependent Chinese Restaurant Process model \cite{Yang-etal2015}. The model clusters event mentions based on their relative distances, given by a learnable distance function. This is the most recent coreference system to be evaluated on the ECB+ corpus. 
\item Iterative WD/CD Classifier: This baseline is produced by the iterative event coreference model that gradually builds event clusters by exploiting inter-dependencies within both WD and CD mentions until the clusters converge  \cite{Choubey-etal2017}.
\end{enumerate}
\subsection{Our System}
We evaluate our system with two sets of slightly different testing
data. The first set uses the gold standard event mentions marked in
the ECB+ corpus (WD-gold and CD-gold);  the second set uses events
marked using the aforementioned event detection tool (WD-detect and
CD-detect).  We can also compare the results from the latter set to  the
last two baseline systems mentioned above. 

\subsection{Results}
Table \ref{wdresults} shows the results obtained for WD
coreference while Table \ref{cdresults} shows the results
obtained for CD coreference. We notice that in the case of WD
coreference, our system overall performs  slightly better than the
state-of-the-art \cite{Choubey-etal2017} for the $CoNLL$ $F_1$ score.  However for CD coreference
our system performs more than 8\% points better than the
state-of-the-art in the same measure. Additionally our WD and CD
systems perform better than the Lemma baseline for all measures except
for $MUC$. There are instances where our WD and CD systems perform
better than the WD-gold and CD-gold, this could be attributed to the
fact that we select all predicted mentions as long as they are part of
the gold standard, a limitation that is needed because of the
incompletely marked ECB+ corpus.  We also notice that our WD-gold and
CD-gold systems perform, as expected, significantly better than the
WD-detect and CD-detect systems. These numbers provide an indication
of how our system performs with a fully annotated ECB+ corpus and an
event detection tool that has 100\% recall.  

\begin{table*}[t!]
\begin{center}
\resizebox{\linewidth}{!}{%
\begin{tabular}{|l|c|c|c|c|c|c|c|c|c|c|}
\hline
\multirow{3}{*}{} & \multicolumn{10}{c|}{\textbf{WD Model}} \\ \cline{2-11} 
 & \multicolumn{3}{c|}{\textbf{$MUC$}} & \multicolumn{3}{c|}{\textbf{$B^3$}} & \multicolumn{3}{c|}{\textbf{$CEAF_e$}} & \multicolumn{1}{c|}{\textbf{$CoNLL$}} \\ \cline{2-11} 
 & \textbf{$R$} & \textbf{$P$} & \textbf{$F_1$} & \textbf{$R$} & \textbf{$P$} & \textbf{$F_1$} & \textbf{$R$} & \textbf{$P$} & \textbf{$F_1$} & \textbf{$F_1$} \\ \hline
Baseline 1: Lemma, Yang et al., 2015& 56.80 & \textbf{80.90} & \textbf{66.70} & 35.90 & 76.20 & 48.80 & 67.40 & 62.90 & 65.10 & 60.20 \\ \hline
Baseline 2: Yang et al., 2015 & 41.70 & 74.30 & 53.40 & 67.30 & 85.60 & 75.40 & \textbf{79.80} & 65.10 & 71.70 & 66.83 \\ \hline
Baseline 3: Choubey et al., 2017 & \textbf{58.50} & 67.30 & 62.60 & \textbf{69.20} & 76.00 &
                                                                    72.40
              & 67.90 & 76.10 & 71.80 & 68.93 \\ \hline \hline
WD-detect (this paper) & 48.16 & 72.75 & 57.95 & 66.10 & \textbf{90.53} & \textbf{76.41} & 68.38 & \textbf{80.02} & \textbf{73.74} & \textbf{69.37} \\ \hline  \hline
WD-gold (this paper) & 67.68 & 71.57 & 69.57 & 85.54 & 87.99 & 86.75 & 80.63 & 79.70 & 80.16 & 78.83 \\ \hline
\end{tabular}
}
\end{center}
\caption{WD Coreference Results}
\label{wdresults}
\end{table*}

\begin{table*}[t!]
\begin{center}
\resizebox{\linewidth}{!}{%
\begin{tabular}{|l|c|c|c|c|c|c|c|c|c|c|}
\hline
\multirow{3}{*}{} & \multicolumn{10}{c|}{\textbf{CD Model}} \\ \cline{2-11} 
 & \multicolumn{3}{c|}{\textbf{$MUC$}} & \multicolumn{3}{c|}{\textbf{$B^3$}} & \multicolumn{3}{c|}{\textbf{$CEAF_e$}} & \multicolumn{1}{c|}{\textbf{$CoNLL$}} \\ \cline{2-11} 
 & \textbf{$R$} & \textbf{$P$} & \textbf{$F_1$} & \textbf{$R$} & \textbf{$P$} & \textbf{$F_1$} & \textbf{$R$} & \textbf{$P$} & \textbf{$F_1$} & \textbf{$F_1$} \\ \hline

Baseline 1: Lemma,  Yang et al., & 39.50 & 73.90 & 51.40 & 58.10 & 78.20 & 66.70 & 58.90 & 36.50 & 46.20 & 54.80 \\ \hline
Baseline 2: Yang et al., 2015 & 67.10 & 80.30 & 73.10 & 40.60 & 78.50 & 53.50 & 68.90 & 38.60 & 49.50 & 58.70 \\ \hline
Baseline 3: Choubey et al., 2017 &\textbf{ 67.50} & \textbf{80.40} & \textbf{73.40} & 56.20 & 66.60 &
                                                                      61.00
              & 59.00 & 54.20 & 56.50 & 63.63 \\ \hline \hline
CD-detect (this paper) & 53.94 &62.09 & 57.73 & \textbf{82.18} & \textbf{84.86} & \textbf{83.50} & \textbf{77.36} & \textbf{72.27} & \textbf{74.73} & \textbf{71.99}\\ \hline \hline
CD-gold (this paper) & 76.09 & 61.24 & 67.86 & 90.15 & 75.68 & 82.28 & 70.10 & 80.58 & 74.98 & 75.04 \\ \hline
\end{tabular}
}
\end{center}
\caption{CD Coreference Results}
\label{cdresults}
\end{table*}

\section{Discussion} 
\subsection{Error Analysis}

Since we construct coreferent clusters of event mentions in an agglomerative way, the primary disadvantage we face is error propagation. To mitigate this, we employ  high thresholds to determine if two clusters (components) can be merged. However, there still are coreferent event pairs that are not considered coreferent and non-coreferent event pairs that are considered coreferent. We have performed an analysis of our system's final predictions on the development set to identify why we get these errors. 
\begin{itemize}
\item \textbf{Missed coreference links:} This issue is common among
  event pairs that are not recognized by the word2vec model as being
  similar to each other, hence causing low relational similarity
  scores. For example, consider event pairs like (\textit{raid},
  \textit{heist}),  (\textit{sprained},
  \textit{injury}), 
  and 
  (\textit{nab}, \textit{grabbing}). 
  These pairs have low relational feature similarities
  causing the classifier to give low coreference scores. Another issue is
  that the
  current word2vec model is not able to successfully compare between
  multi-word events like \textit{scooping up}, \textit{making off},
  \textit{cleaning up} and \textit{stay alive}. Since our model uses
  only event information in order to determine a coreference score,
  we also miss out on cases where event-entity coreference becomes
  crucial. Consider the following pair of sentences: 
\begin{quote}
{\small \textit{Pierre Thomas was \textbf{placed} on injured reserve by the New Orleans Saints on Wednesday, meaning he won't play in the 2011 NFL playoffs.}}
\end{quote}
and
\begin{quote}
{\small \textit{This means they will be missing two of their best players in the rushing game, and \textbf{it} could weaken the attack that the Saints have on offense even further.}}
\end{quote}
In this case, having entity coreference information, and also event arguments and relations will help better resolve the coreference.
\item \textbf{Incorrect coreference Links:} This issue is common among
  event pairs within the same sentence. Since our system uses
  contextual features, events in the same sentence share similar
  contextual features, resulting in our NNs giving high coreference
  scores to these event pairs. For example, consider the sentence
  paired with itself, where we are testing for a coreference within
  the same sentence.
\begin{quote}
{\small \textit{HP today announced that it has signed a definitive
    agreement to acquire EYP Mission Critical Facilities Inc., a
    consulting company specializing in strategic technology
    \textbf{planning}, design, and operations support for large-scale data centers.}}
\end{quote}
and
\begin{quote}
{\small \textit{HP today announced that it has signed a definitive agreement to acquire EYP Mission Critical Facilities Inc., a consulting company specializing in strategic technology planning, \textbf{design}, and operations support for large-scale data centers.}}
\end{quote}

Here the contextual similarity between the two events \emph{planning}
and \emph{design} are high, being very close to each other.

\end{itemize}

\section{Conclusion and Future Work}

Our results show that  the pairwise models and the  method with which we find coreferent clusters of event mentions is adept at identifying event coreference even without relying on event arguments. These results make manifest that accurate event detection significantly helps in improving event coreference resolution, as evidenced by the disparity between our system results when event mentions detected are gold standard and otherwise. 

We would like to explore if event arguments and their relations to the
event mention (whether semantic or syntactic) can be extracted in a
reasonable way without error which would propagate upwards. We would
also want to explore using the GloVe vector model embeddings \cite{pennington-socher-manning:2014:EMNLP2014} in our NNs to obtain
better word sense disambiguation, which would help in reducing the
number of coreferent event pairs that are not recognized currently. We
would also like to do joint entity and event coreference to
improve the overall event coreference resolution. Additionally, we
currently employ an even distribution of coreferent and non-coreferent
event pairs for training purposes. However, since there is a much
larger number of non-coreferent event pairs in general, we would also
like to explore how our system performs with a higher number of
non-coreferent event pairs for training.

\bibliographystyle{acl.bst}
\bibliography{main}

\end{document}